\documentclass[10pt,twocolumn,letterpaper]{article}

\usepackage{iccv}
\usepackage{times}
\usepackage{epsfig}
\usepackage{graphicx}
\usepackage{amsmath}
\usepackage{amssymb}
\usepackage{bbm}
\usepackage{algorithm}  
\usepackage{algorithmicx}
\usepackage{algpseudocode}
\usepackage{multirow}
\usepackage{arydshln}
\usepackage[switch]{lineno}
\usepackage{booktabs}
\usepackage{adjustbox}

\usepackage[breaklinks=true,bookmarks=false,backref=page]{hyperref}
\hypersetup{
    colorlinks=true,
    linkcolor=blue,
    filecolor=magenta,      
    urlcolor=cyan,
    pdftitle={Overleaf Example},
    pdfpagemode=FullScreen,
    }

\iccvfinalcopy 

\def\iccvPaperID{****} 

\ificcvfinal\fi

\begin{document}

\title{Online Continual Learning via Multiple Deep Metric Learning and Uncertainty-guided Episodic Memory Replay \\ 3\textsuperscript{rd} Place Solution for ICCV 2021 Workshop SSLAD Track 3A Continual Object Classification}

\author{Muhammad Rifki Kurniawan\\
Xi'an Jiaotong University\\
China\\
{\tt\small mrifkikurniawan17@gmail.com}
\and
Xing Wei\\
Xi'an Jiaotong University\\
China\\
{\tt\small weixing@mail.xjtu.edu.cn}
\and
Yihong Gong\\
Xi'an Jiaotong University\\
China\\
{\tt\small ygong@mail.xjtu.edu.cn}
}

\maketitle
\ificcvfinal\thispagestyle{empty}\fi

\begin{abstract}
    Online continual learning in the wild is a very difficult task in machine learning. Non-stationarity in online continual learning potentially brings about catastrophic forgetting in neural networks. Specifically, online continual learning for autonomous driving with SODA10M dataset exhibits extra problems on extremely long-tailed distribution with continuous distribution shift. To address these problems, we propose multiple deep metric representation learning via both contrastive and supervised contrastive learning alongside soft labels distillation to improve model generalization. Moreover, we exploit modified class-balanced focal loss for sensitive penalization in class imbalanced and hard-easy samples. We also store some samples under guidance of uncertainty metric for rehearsal and perform online and periodical memory updates. Our proposed method achieves considerable generalization with average mean class accuracy (AMCA) 64.01\% on validation and 64.53\% AMCA on test set. Code is available at: \href{https://github.com/mrifkikurniawan/sslad}{https://github.com/mrifkikurniawan/sslad}
\end{abstract} 

\section{Introduction}
Training neural networks in continual fashion in the wild environment is a extremely challenging task in deep learning (DL). The existing DL systems particularly rely on the strong assumption of independent and identically distributed or i.i.d sampling from the dataset and minimize the loss with respect to those samples. However, those would suffer from catastrophic forgetting while updating the networks on non-stationary data when na\"ive minimization on cross entropy loss is employed in classification task. In order to alleviate this issue, the available techniques span from moderating the non-i.i.d setting via replaying the past data points or called rehearsal-based approach \cite{Rebuffi2017iCaRLIC, LopezPaz2017GradientEM}, constraining the parameters update or regularization \cite{Dong_Hong_Tao_Chang_Wei_Gong_2021, Kirkpatrick2017OvercomingCF}, or introducing architectural modification approach \cite{Rusu2016ProgressiveNN, yoon2018lifelong}.

\begin{figure}[!t]
	\centering
	\includegraphics[width=0.9\linewidth]{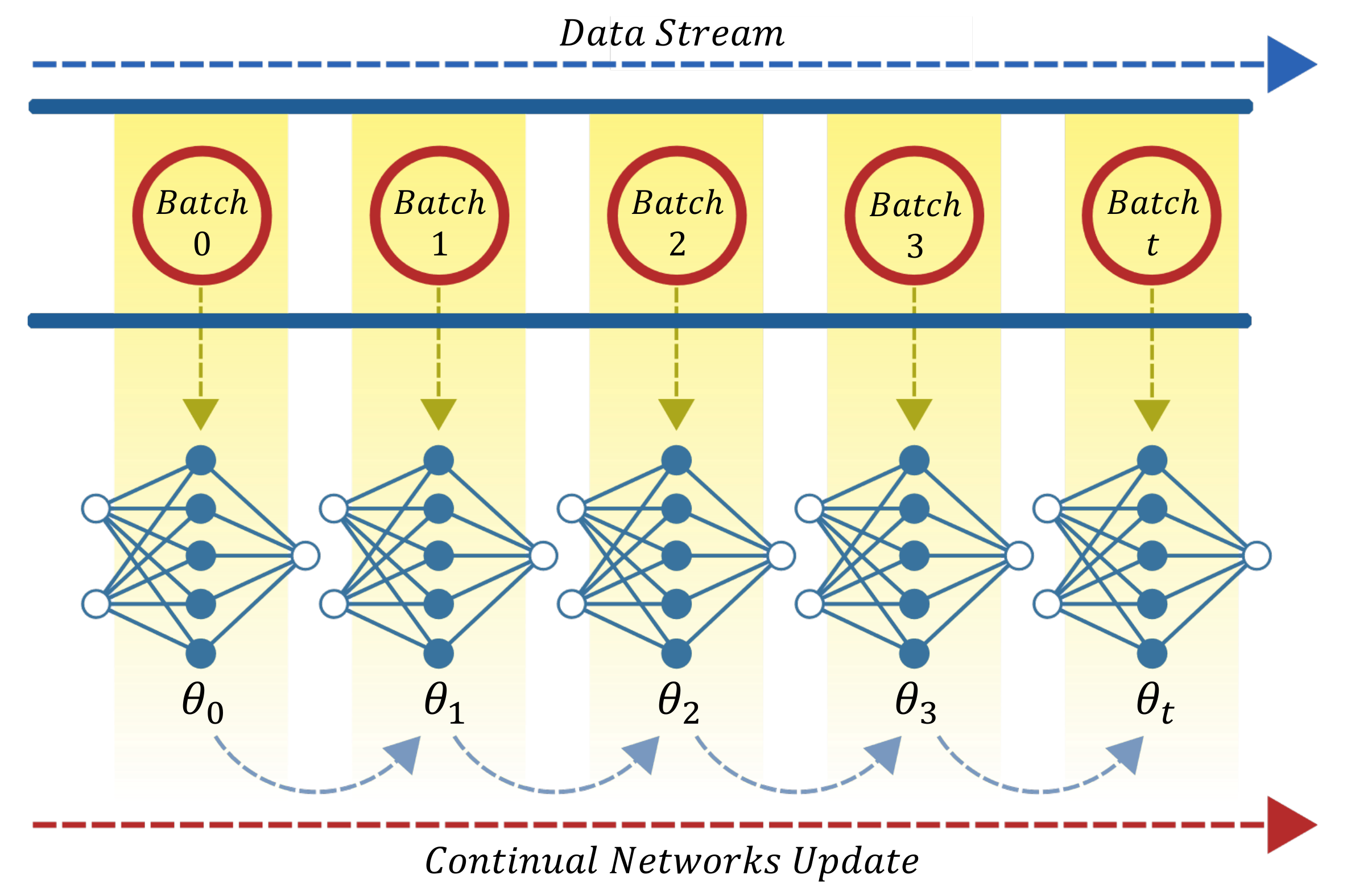}
	\caption{The overall framework of online continual learning where the inter-tasks boundaries are inaccessible and model parameters $\theta$ are continuously updated.}
	\label{fig:online_cl}
\end{figure}

This Self-supervised Learning for Next-Generation Industry-level Autonomous Driving (SSLAD) workshop track 3-A challenge scenario presents online continual learning with relatively more realistic world setup. Different from common class incremental learning, this challenge established fixed predefined classes instead of incrementally novel classes and the distributions are changing over time in continuously streaming dataset as depicted on Fig. \ref{fig:online_cl}. Therefore, the vanilla deep learning classification with gradient descent and cross-entropy loss is vulnerable to suboptimal solutions and results in forgetting catastrophically. In addition, as the data points are incoming in real-time, the inter-tasks boundaries are unavailable resulting in inability to revisit certain task dataset. And the coming batch may be probably extremely imbalanced between classes that leads to long-tail distribution. 

\begin{figure*}[!t]
	\centering
	\includegraphics[width=0.9\linewidth]{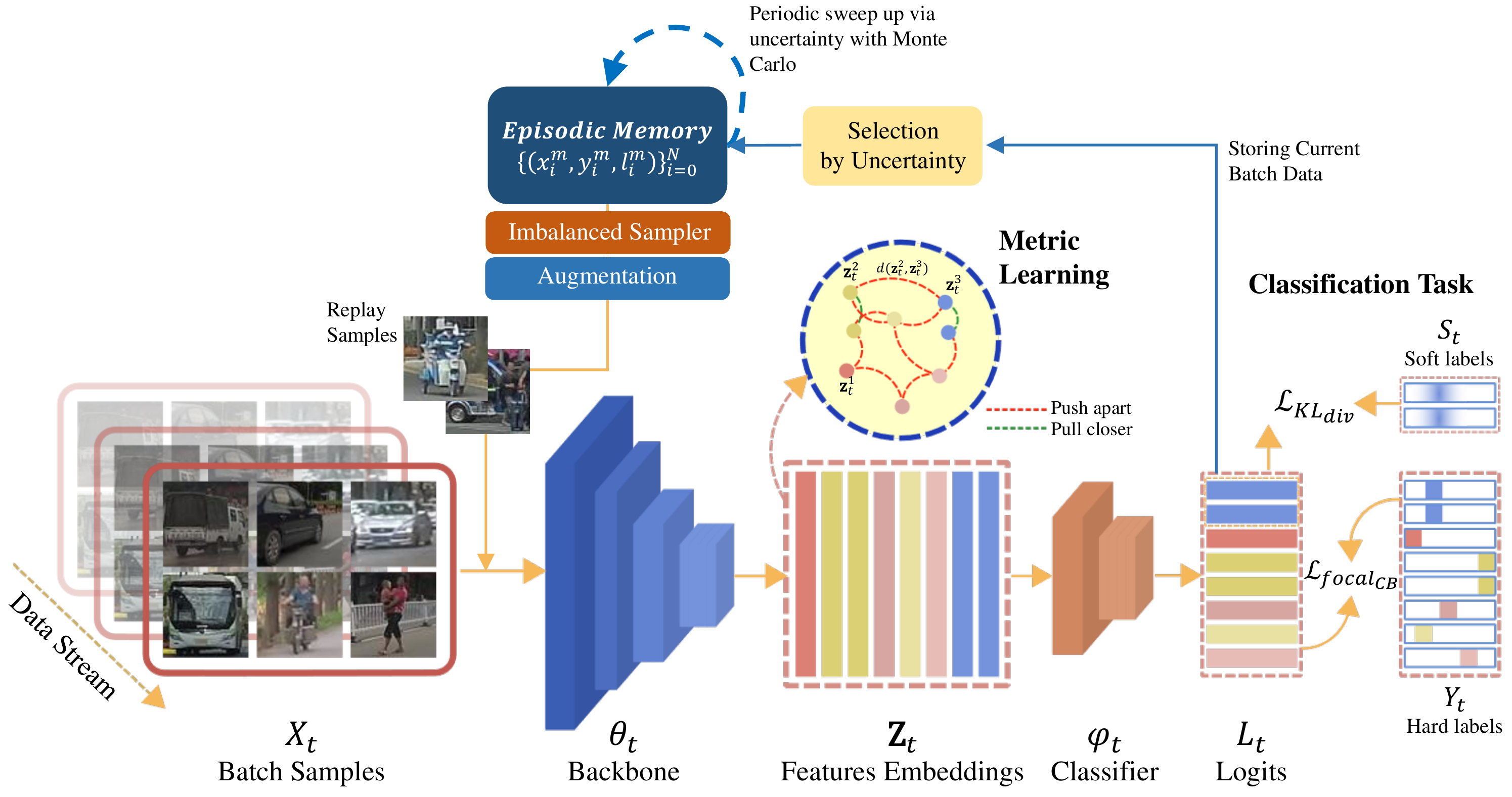}
	\caption{Our proposed learning method where we train the model representations via metric similarity learning with contrastive and supervised contrastive learning. Moreover, we recollect past logits to generate soft labels as the networks targets alongside one-hot labels to train classification task. Our memory is updated periodically and continually based on uncertainty measure.}
	\label{fig:architecture}
\end{figure*}

Considering those demanding issues, we proposed multiple deep metric learning in order to build generalized representation learning along with soft labels distillation and replay method with uncertainty-based progressive episodic memory. Seeing that in online setting we should select the samples among batch inputs at a time, we employ uncertainty-aware sampling strategy with entropy measure as the metric for selection. As a result, given a batch of samples in a stream, the sampler will select the top-$k$ most uncertain samples with highest entropy to be stored in episodic memory. Moreover, we executed periodic memory sweeps up via uncertainty approximation with Monte-Carlo method as proposed by Rainbow Memory \cite{Bang_2021_CVPR} and keep the top-$k$ most uncertain while detach the rest.

Generally, our proposed method for this competition cover
\begin{itemize}
  \item Multiple task-agnostic deep metric representation learning using both contrastive and supervised contrastive learning for improving model generalization.
  \item Online and periodic sampling strategy under guidance of uncertainty measure for replay buffer.
  \item Classification task learning with soft labels retrospection and modified class balanced focal loss for addressing long-tail recognition.
\end{itemize}

\section{Proposed Method}
Our proposed method ranges from employing multiple deep metric learning via contrastive and supervised contrastive learning to building task-agnostic representation for alleviating distribution shift. While, the problem of forgetting was reduced by replaying the sorted-by-uncertainty samples in both online and occasional ways. Since the batch class distribution is potentially imbalanced, we introduce modified class-balanced focal loss for the classification task.

\subsection{Multiple Metric Learning}
In this section, we introduce the generalized representation learning with task agnostic deep metric learning. In particular, we employ and get together the benefits of multiple similarity metric representation learning using contrastive \cite{1467314} and supervised contrastive learning \cite{NEURIPS2020_d89a66c7}. Different from contrastive learning where mainly based on pairwise one positive per anchor, the supervised contrastive learning exploits a cluster of positive points per anchor obtained from the labels. We have experimented that combining both losses with patiently tuned hyperparameters assigned to each loss improves the accuracy significantly.

Given the current vector embeddings $\mathbf{Z}_t$ on batch $t$ extracted from backbone $\theta_t$ and their associated labels $Y_t$, we pass those into both metric loss as follows

\begin{equation}\label{deep-metric-loss}
    \mathcal{L}_{dml} = \alpha\mathcal{L}_{Con}\left (\mathbf{Z}_t, Y_t  \right ) + \beta\mathcal{L}_{SupCon}\left ( \mathbf{Z}_t, Y_t\right ).
\end{equation}
Where $\mathcal{L}_{dml}$ denotes the total deep metric loss, $\mathcal{L}_{Con}$ and $\mathcal{L}_{SupCon}$ are the contrastive and supervised contrastive loss, respectively. We add them together with hyperparameters $\alpha$ and $\beta$ for each individual loss to control each loss strength to the total loss. We set $\alpha$ and $\beta$ with value 0.3 and 0.1 consecutively obtained from grid search. Unlike the original implementation of supervised contrastive learning \cite{NEURIPS2020_d89a66c7}, we did not project the embeddings from backbone to lower dimensional space rather maintain the original embedding dimension.

\subsection{Classification with Retrospection and Class-balanced Focal Loss}
Aiming improving generalization, avoiding overfitting, and reducing catastrophic forgetting, we were also smoothing the targets via retrospecting the logits $l^m_i$ of the past stored in memory. Inspired by \cite{Deng_Zhang_2021}, we construct soft label $s_i$ from image $i$ by converting logits with normalized softmax with temperature as the Eq. (\ref{softlabels-retros}) below. 

\begin{equation} \label{softlabels-retros}
    s_i = \sigma \left ( l^m_i/\tau \right ) = \frac{\mathrm{exp}\left ( l^m_i/\tau \right )}{\sum _j \mathrm{exp}\left ( l^m_{i,j}/\tau \right )}. 
\end{equation}
Which $\sigma$ indicates softmax function, $\tau$ is the temperature that is configured as 0.5, and $j$ is the number of classes. 

Instead of using cross-entropy as the loss function, we proposed focal loss \cite{Lin2017FocalLF} with class-balanced modification in Eq. (\ref{class-balanced-fl}) for this continual setting since the incoming batch dataset might be extremely imbalanced. The focal loss penalizes higher for the difficult samples and less sensitivity to easier samples. Meanwhile, the auxiliary class-balance weighting will penalize greater for underrepresented classes and vice versa. 

\begin{equation} \label{class-balanced-fl}
    \mathcal{L}_{focal_{CB}} = -\alpha\frac{1-\beta}{1-\beta^{n_y}}\left ( 1-p_i \right )^{\gamma }\mathrm{log}\left ( p_i \right )y_i.
\end{equation}
We use $\alpha$-balanced focal loss variant where adjusted at 0.5 and $\gamma$ adjusted to 0.0, $\beta$ is the class-balanced hyperparameter which set to 0.81, $n_y$ is the classes distribution in the current batch, $p_i$ denotes probability distribution and $y_i$ is the class labels. 

For the classification task, we sum up the class-balanced focal loss for hard labels targets and compute the discrepancy between retrospected soft labels and probability distributions using Kullback–Leibler divergence loss in Eq. (\ref{classification-loss}). We control both loss with hyperparameters $\gamma$ and $\delta$ for focal loss and KL loss respectively. We set the $\gamma$ to 1. Meanwhile, since the soft labels quality depend on the model $\omega_t$ and those should be improved as the model learns, we progressively tune up the hyperparameter $\delta$ in pre-defined milestones.
\begin{equation}\label{classification-loss}
    \mathcal{L}_{\mathrm{cls}} = \gamma\mathcal{L}_{focal_{CB}}\left ( p_i, y_i \right ) + \delta\mathcal{L}_{KL_{div}}\left ( f_{\omega_t}\left ( x^m_i \right ),s_i \right ).
\end{equation}

\subsection{Uncertainty-aware Replay}
In order to address the forgetting problem, we stored some samples into the episodic memory and then replayed them with the current batch stream. Our method applied online and periodical memory update. The online update aims to select the current batch samples into memory while the periodic sampling re-sorting and sweeping the current memory samples. The underlying measure for memory management in our method is basically the uncertainty value which is assigned to each single data point. 

\textbf{Online Memory Update}. we proposed a sampling scoring method applied to each single data point by computing both entropy and considering hard negative mining. Given $i$ samples within a batch, the sampling score of each of them will be computed with Eq. (\ref{uncertainty-scoring}) below. 
\begin{equation} \label{uncertainty-scoring}
\begin{gathered}
u_i = - \sum_{k=0}^{C} p_k \log_e\left ( p_k \right ) + n_s, \\ 
\mathrm{where} \; n_s :=
\left\{
	\begin{array}{ll}
		0.5  & \mbox{if } \hat{y}_i\neq y_i \\
		0 & \mathrm{else.}
	\end{array}
\right.
\end{gathered}
\end{equation}
Here, $u_i$ is the sampling score, $p_k$ is the probability of class $k$, $C$ denotes the number of classes, and $n_s$ indicates predefined hard negative score. Later, we sort the current batch samples with this score $s$ with descending order and store top-5 samples based on this proposed score.  

\begin{algorithm}[!t]
\caption{Our proposed online memory update.} 
\label{online_sampler}
{\bf Input:} 
$K$ denotes batch size, $S$ denotes number of selected samples, $\mathfrak{D}^M_t$ denotes current memory, $\mathfrak{D}_t = \left \{ \left ( x_i, y_i \right ) \right \}_{i=0}^{K}$ denotes current batch inputs, $C$ denotes maximum memory capacity.\\
{\bf Output:}
Updated memory $\mathfrak{D}^M_{t+1}$ \\
\begin{algorithmic}[1]
\State Sort $\mathfrak{D}_t$ by $u$ computed by Eq. (\ref{uncertainty-scoring}) \Comment\textcolor{blue}{sort descendingly}
\State $C_t \leftarrow \mathrm{len}\left ( \mathfrak{D}^M_t \right )$ \Comment\textcolor{blue}{get current capacity}
\If{$C-C_t \geq S $}
    \State  {$\mathfrak{D}^M_{t+1} \leftarrow \mathfrak{D}_t^M \cup  \mathfrak{D}_t$[0:$S$]}
\Else
    \State  $\mathfrak{D}^M_{t+1} \leftarrow \mathfrak{D}_t^M \cup  \mathfrak{D}_t$[0:$C-C_t$]
\EndIf
\end{algorithmic}
\end{algorithm}

\textbf{Periodic Memory Update}. Since stream data is continuously arriving and the memory buffer is growing up while the capacity remains fixed, we execute periodic updates to sweep up most certain samples within memory. Instead of forwarding the input in a single pass, we adopt Rainbow Memory \cite{Bang_2021_CVPR} strategy that estimates the uncertainty via monte-carlo with multiple passes forward as in Eq. (\ref{rm}) given some input $x_i$ with some $T$ augmentations.
\begin{equation} \label{rm}
\begin{gathered}
S_c = \sum_{t=1}^T \mathbbm{1}_c \operatorname*{argmax}_{\hat{c}}\: p\left ( y=\hat{c}\mid \tilde{x}_i \right )  
, \\ 
u\left ( x \right ) = 1 - \frac{1}{T}\operatorname*{max}_{c}S_c
\end{gathered}
\end{equation}
Where, $T$ is the number of transformations, $c$ denotes the number of classes, and $x_i$ is the input image $i$, $S_c$ is the count of the model predicting a certain class, while $u(x)$ is the uncertainty score of sample $x$. As the model predicts inconsistently given some perturbed samples from the same image, the uncertainty score will be higher. Finally, our method will keep top-500 samples with highest uncertainty and subsequently replaying these most uncertain samples first in an online setting.

\begin{algorithm}[!t]
\caption{Our proposed periodic memory update.} 
\label{algo:periodic_sampler}
{\bf Input:} 
$K$ denotes batch size, $S$ denotes default memory size after reset, $\mathfrak{D}^M_t$ denotes current memory, $\mathfrak{D}_t = \left \{ \left ( x_i, y_i \right ) \right \}_{i=0}^{K}$ denotes current batch inputs, $C$ denotes maximum memory capacity.\\
{\bf Output:}
Updated memory $\mathfrak{D}^M_{t+1}$ \\
\begin{algorithmic}[1]
    \State $\mathfrak{D}^M_t \leftarrow \mathfrak{D}^M_t \cup \mathfrak{D}_t$ 
    \State Sort $\mathfrak{D}^M_t$ by $u\left ( x \right )$ computed by Eq. (\ref{rm})
    \State  $\mathfrak{D}^M_{t+1} \leftarrow \mathfrak{D}_t^M$[0:$S$]
\end{algorithmic}
\end{algorithm}

\subsection{Our Networks Learning Algorithm}
This subsection wraps our proposed method from input to parameters update as illustrated in Fig. \ref{fig:architecture}. Given the dataset stream $\mathfrak{D}_s$=$\left \{ \left ( X_t, Y_t \right ) \right \}_{t=0}^{\infty}$ where is boundless here, our method joined current batch input $X_t$, $Y_t$ with the augmented episodic memory sample containing triplets of images $X^m_t$, labels $Y^m_t$, and logits $L^m_t$ selected using imbalanced-aware sampling\footnote{\href{https://github.com/ufoym/imbalanced-dataset-sampler}{https://github.com/ufoym/imbalanced-dataset-sampler}}. The feature embeddings extracted from the backbone are then leveraged to measure the similarity distance between them in the metric learning module. At the head of architecture, we train networks with both hard labels and the soft labels distillation with associated class-balanced focal loss and KL divergence loss, consecutively. Finally the method updates the memory in the online setting using algorithm \ref{online_sampler} and periodically clears the memory with algorithm \ref{algo:periodic_sampler}. 

The complete training procedure is summarized in algorithm \ref{train_algorithm}. As shown in the algorithm \ref{train_algorithm}, while updating the networks parameters, the gradients from the classification loss are accumulated with metric learning gradients to do single step parameters update. Our training algorithm requires only single step forward and backward for updating the model parameters and episodic memory. Thus maintain real-time updates which should be applicable in real-time online continual scenario.

\begin{algorithm}[!t]
\caption{Our proposed training scheme.} 
\label{train_algorithm}
{\bf Input:} 
Networks backbone $\theta_t$, head $\varphi_t$, and full-networks $\omega_t$, episodic memory $\mathfrak{D}^M_t$, max memory capacity $C$, learning rate $\eta$, imbalanced data sampler $\textup{S}_{ids}$.\\
{\bf Output:}
Trained model $\omega$\\
\begin{algorithmic}[1]
\For{$X_t$, $Y_t$ in $\mathbf{X}_{stream}$: } 
    \State $X^m_t, Y^m_t, L^m_t \leftarrow \textup{S}_{ids}\left ( \mathfrak{D}^M_t \right )$\Comment\textcolor{blue}{replay samples}
    \State $X_t, Y_t \leftarrow X_t \cup \mathrm{aug}\left ( X^m_t \right ), Y_t \cup Y^m_t$
    \State $\mathbf{Z}_t \leftarrow f_{\theta}\left ( X_t \right )$\Comment\textcolor{blue}{get embeddings}
    \State $L_t \leftarrow f_{\varphi}\left ( \mathbf{Z}_t \right )$\Comment\textcolor{blue}{get outputs logits}
    \State $\mathcal{L_{\mathrm{dml}}} \leftarrow$ Compute loss with equation (\ref{deep-metric-loss}).
    \State $\mathcal{L}_{\mathrm{cls}} \leftarrow$ Compute loss with equation (\ref{classification-loss}).
    \State $\theta_{t+1} \leftarrow \theta_t - \eta \nabla_{\theta}\mathcal{L}_{\mathrm{dml}}$\Comment\textcolor{blue}{Update backbone $\theta$}
    \State $\omega_{t+1} \leftarrow \omega_t - \eta \nabla_{\omega}\mathcal{L}_{\mathrm{cls}}$\Comment\textcolor{blue}{Update full networks $\omega$}
    \State $\mathfrak{D}^M_{t+1} \leftarrow$ update memory by algorithm (\ref{online_sampler}),
    \If{$\mathrm{len}\left ( \mathfrak{D}^M_{t+1} \right )= C$ or $t\%1000=0$}
        \State $\mathfrak{D}^M_{t+1}\leftarrow$ update memory with algorithm (\ref{algo:periodic_sampler}).
    \EndIf
\EndFor
\end{algorithmic}
\end{algorithm}

\section{Experiments and Results}
\subsection{Datasets}
This competition utilizes SODA10M \cite{han2021soda10m} dataset containing large-scale 2D object detection for autonomous driving for self-supervised and semi-supervised learning task exploration. The dataset mainly contains 10M unlabeled images and 20K labeled images splitted into training (5K), validation (5K) and test (10K) sets. Since this challenge scenario is the supervised setting, the unlabeled data are not used here. This dataset comprised 6 classes of pedestrian, cyclist, car, truck, tram, and tricycle with heavily long-tailed distribution which is predominated by car and pedestrian class. Due to solve classification task, the original full-frame images are cropped into instance-level.
\subsection{Implementation Details}
\textbf{Architecture}. Our proposed method was built on top of ResNet50-D \cite{He2019BagOT} with pretrained weight from ImageNet \cite{Deng2009ImageNetAL}. We find that naïve fully-connected networks without bias perform better than with bias or multi-layer perceptrons. The networks input images are preserved in original size in 64x64 resolution.

\textbf{Training strategy}. Our method tunes the parameters update via Stochastic Gradient Descent (SGD) with initial learning rate 0.0111 obtained from grid search. We were also reconfiguring the learning rate after 1000 and 2000 iterations with factor 0.1. We set the input stream batch size to 10 as the maximum allowed size and replayed 6 batch size augmented memory exemplar. This method was trained on top of Avalanche \cite{lomonaco2021avalanche} framework along with log every experiment results with Weights \& Biases \cite{wandb}. We train the networks on single GPU NVIDIA GeForce Titan XP 12 GB.

\textbf{Episodic Memory and Replay}. We set up the memory maximum capacity to 1000 objects. Our method also stores the top-5 highest sampling score based on Eq. (\ref{uncertainty-scoring}) in online scenario. Meanwhile, periodically sweep up memory either every 1000 iterations or if the memory is getting maximum with Eq. (\ref{rm}). While training, we select 6 samples for each iteration to be joined with current stream data and, subsequently, augment those with random horizontal flip, random rotation, and color jitter.  

\subsection{Results and Discussion}

\begin{table}[!tb]
\small
\centering
\begin{adjustbox}{width=\columnwidth,center}
\begin{tabular}{cccccc}
\toprule
\multirow{1}{*}{Method} & \multirow{1}{*}{Val AMCA} & \multirow{1}{*}{Test AMCA} \\
\midrule
Baseline (Uncertainty Replay)$^*$ & 57.517  & -    \\
+ Multi-step Lr Scheduler$^*$ & 59.591 (+2.07)  & -   \\
+ Soft Labels Retrospection$^*$ & 59.825 (+0.23)  & -  \\
+ Contrastive Learning$^*$ & 60.363 (+0.53) & 59.68  \\
+ Supervised Contrastive Learning$^*$ & 61.49 (+1.13)  & -  \\
+ Change backbone to ResNet50-D$^*$ & 62.514 (+1.02)  & -  \\
+ Focal loss$^*$  & 62.71 (+0.19)  & -  \\
+ Cost Sensitive Cross Entropy & 63.33 (+0.62)  & -  \\
+ Class Balanced Focal loss$^*$  & \textbf{64.01 (+1.03)}  & \textbf{64.53 (+4.85)} \\
+ Head Fine-tuning with \\ Class Balanced Replay & \textcolor{red}{65.291 (+1.28)}  & \textcolor{red}{62.58 (-1.56)}  \\
+ Head Fine-tuning with \\ Soft Labels Retrospection & \textcolor{red}{66.116 (+0.83)} & \textcolor{red}{62.97 (+0.39)}  \\
\bottomrule
\end{tabular}
\end{adjustbox}
\raggedright\footnotesize $^*$Applied to our final method. \newline
\caption{Our proposed method benchmark with its associated tricks and improvements on both validation and test set in average mean class accuracy (AMCA) metric.}
\label{tab:results}
\end{table}

Generally our strategies to address this SSLAD challenge and boost up the accuracy consider constructing general representation learning that performs well in any distributions. We assume that there exist network parameters in parameters space that have low loss landscape in many distributions and pose wider loss curvature. That is why our tricks are mainly thinking of any trick related to generalization improvement. Overall, the results of our experiments have been summarized in table \ref{tab:results}. Based on the experiments, understanding when to re-setup learning rate in continual learning is essential. Reconfiguring the learning rate using multi-step learning rate boosts up performance the most with 2.07\% increasing in average mean class accuracy (AMCA). Large learning rate leads to fast adaptation beneficial for the beginning of the training stage from ImageNet features to targeted domain but most vulnerable to catastrophic forgetting. Hence, slowing down the learning rate could alleviate the trained model from massive forgetting the past but remain reasonably plastic.

Meanwhile, networks training using contrastive learning with class balanced focal loss and soft labels retrospection performs best in maintaining the overfit-generalization trade-off which is able to get comparable accuracy between validation and test set, even getting higher on test set. Multiple deep metric learning contributes second greatest to the accuracy increasing with 1.66\% boost up confirming the importance of auxiliary representation learning. Additionally, we are also exploring class imbalanced strategy via modified weighted loss including cost sensitive and class-balanced weighting. The result shows that class balanced weighting associated with focal loss performs best in our experiments with larger gap improvement than cost sensitive cross entropy. This indicates that taking into account hard-easy samples and batch-wise imbalance distribution to loss in this domain is really helpful for accuracy.

Actually, since the most vulnerable to forgetting and crucial to make decisions is the classifier head, we attempted to calibrate the classifier head via fine-tuning with class balanced replay. By this we rehearse samples with classes balanced distribution to calibrate head such that the predicted outputs are unbiased to certain classes. As we can see in the table \ref{tab:results}, this trick improves validation AMCA but leads to overfitting and significant performance drop on the test set. We expect bag of tricks related to generalization improvement for head classifier including embedding replay with augmentation or generative latent replay should be helpful.

\section{Conclusion}

In this paper, we proposed generalized task-agnostic representation learning via multiple deep metric learning with contrastive and supervised contrastive learning along with soft labels retrospection for online continual learning. Rather than naive cross entropy loss for solving classification task, we developed class-balanced focal loss for training the networks to do classification. Moreover, we stored some samples to episodic memory in online and periodical manner via entropy-based and rainbow memory sample scoring respectively and replaying some of them in an online setting. Our experiments demonstrate that our proposed method generalizes well on both validation and test set under continuous distribution shift scenario.

{\small
\bibliographystyle{ieee_fullname}
\bibliography{egbib}

\begin{thebibliography}{10}\itemsep=-1pt

\bibitem{Bang_2021_CVPR}
Jihwan Bang, Heesu Kim, YoungJoon Yoo, Jung-Woo Ha, and Jonghyun Choi.
\newblock Rainbow memory: Continual learning with a memory of diverse samples.
\newblock In {\em Proceedings of the IEEE/CVF Conference on Computer Vision and
  Pattern Recognition (CVPR)}, pages 8218--8227, June 2021.

\bibitem{wandb}
Lukas Biewald.
\newblock Experiment tracking with weights and biases, 2020.
\newblock Software available from wandb.com.

\bibitem{1467314}
S. Chopra, R. Hadsell, and Y. LeCun.
\newblock Learning a similarity metric discriminatively, with application to
  face verification.
\newblock In {\em 2005 IEEE Computer Society Conference on Computer Vision and
  Pattern Recognition (CVPR'05)}, volume~1, pages 539--546 vol. 1, 2005.

\bibitem{Deng2009ImageNetAL}
Jia Deng, Wei Dong, Richard Socher, Li-Jia Li, K. Li, and Li Fei-Fei.
\newblock Imagenet: A large-scale hierarchical image database.
\newblock {\em 2009 IEEE Conference on Computer Vision and Pattern
  Recognition}, pages 248--255, 2009.

\bibitem{Deng_Zhang_2021}
Xiang Deng and Zhongfei Zhang.
\newblock Learning with retrospection.
\newblock {\em Proceedings of the AAAI Conference on Artificial Intelligence},
  35(8):7201--7209, May 2021.

\bibitem{Dong_Hong_Tao_Chang_Wei_Gong_2021}
Songlin Dong, Xiaopeng Hong, Xiaoyu Tao, Xinyuan Chang, Xing Wei, and Yihong
  Gong.
\newblock Few-shot class-incremental learning via relation knowledge
  distillation.
\newblock {\em Proceedings of the AAAI Conference on Artificial Intelligence},
  35(2):1255--1263, May 2021.

\bibitem{han2021soda10m}
Jianhua Han, Xiwen Liang, Hang Xu, Kai Chen, Lanqing Hong, Chaoqiang Ye, Wei
  Zhang, Zhenguo Li, Xiaodan Liang, and Chunjing Xu.
\newblock Soda10m: Towards large-scale object detection benchmark for
  autonomous driving, 2021.

\bibitem{He2019BagOT}
Tong He, Zhi Zhang, Hang Zhang, Zhongyue Zhang, Junyuan Xie, and Mu Li.
\newblock Bag of tricks for image classification with convolutional neural
  networks.
\newblock {\em 2019 IEEE/CVF Conference on Computer Vision and Pattern
  Recognition (CVPR)}, pages 558--567, 2019.

\bibitem{NEURIPS2020_d89a66c7}
Prannay Khosla, Piotr Teterwak, Chen Wang, Aaron Sarna, Yonglong Tian, Phillip
  Isola, Aaron Maschinot, Ce Liu, and Dilip Krishnan.
\newblock Supervised contrastive learning.
\newblock In H. Larochelle, M. Ranzato, R. Hadsell, M.~F. Balcan, and H. Lin,
  editors, {\em Advances in Neural Information Processing Systems}, volume~33,
  pages 18661--18673. Curran Associates, Inc., 2020.

\bibitem{Kirkpatrick2017OvercomingCF}
James Kirkpatrick, Razvan Pascanu, Neil~C. Rabinowitz, Joel Veness, Guillaume
  Desjardins, Andrei~A. Rusu, Kieran Milan, John Quan, Tiago Ramalho, Agnieszka
  Grabska-Barwinska, Demis Hassabis, Claudia Clopath, Dharshan Kumaran, and
  Raia Hadsell.
\newblock Overcoming catastrophic forgetting in neural networks.
\newblock {\em Proceedings of the National Academy of Sciences}, 114:3521 --
  3526, 2017.

\bibitem{Lin2017FocalLF}
Tsung-Yi Lin, Priya Goyal, Ross~B. Girshick, Kaiming He, and Piotr Doll{\'a}r.
\newblock Focal loss for dense object detection.
\newblock {\em 2017 IEEE International Conference on Computer Vision (ICCV)},
  pages 2999--3007, 2017.

\bibitem{lomonaco2021avalanche}
Vincenzo Lomonaco, Lorenzo Pellegrini, Andrea Cossu, Antonio Carta, Gabriele
  Graffieti, Tyler~L. Hayes, Matthias~De Lange, Marc Masana, Jary Pomponi, Gido
  van~de Ven, Martin Mundt, Qi She, Keiland Cooper, Jeremy Forest, Eden
  Belouadah, Simone Calderara, German~I. Parisi, Fabio Cuzzolin, Andreas
  Tolias, Simone Scardapane, Luca Antiga, Subutai Amhad, Adrian Popescu,
  Christopher Kanan, Joost van~de Weijer, Tinne Tuytelaars, Davide Bacciu, and
  Davide Maltoni.
\newblock Avalanche: an end-to-end library for continual learning.
\newblock In {\em Proceedings of IEEE Conference on Computer Vision and Pattern
  Recognition}, 2nd Continual Learning in Computer Vision Workshop, 2021.

\bibitem{LopezPaz2017GradientEM}
David Lopez-Paz and Marc'Aurelio Ranzato.
\newblock Gradient episodic memory for continual learning.
\newblock In {\em NIPS}, 2017.

\bibitem{Rebuffi2017iCaRLIC}
Sylvestre-Alvise Rebuffi, Alexander Kolesnikov, G. Sperl, and Christoph~H.
  Lampert.
\newblock icarl: Incremental classifier and representation learning.
\newblock {\em 2017 IEEE Conference on Computer Vision and Pattern Recognition
  (CVPR)}, pages 5533--5542, 2017.

\bibitem{Rusu2016ProgressiveNN}
Andrei~A. Rusu, Neil~C. Rabinowitz, Guillaume Desjardins, Hubert Soyer, James
  Kirkpatrick, Koray Kavukcuoglu, Razvan Pascanu, and Raia Hadsell.
\newblock Progressive neural networks.
\newblock {\em ArXiv}, abs/1606.04671, 2016.

\bibitem{yoon2018lifelong}
Jaehong Yoon, Eunho Yang, Jeongtae Lee, and Sung~Ju Hwang.
\newblock Lifelong learning with dynamically expandable networks.
\newblock In {\em 6th International Conference on Learning Representations,
  {ICLR} 2018, Vancouver, BC, Canada, April 30 - May 3, 2018, Conference Track
  Proceedings}. OpenReview.net, 2018.

\end{thebibliography}
}

\end{document}